\documentclass[10pt]{article} % onecolumn (standard format)

\usepackage[utf8]{inputenc}
\usepackage{amsmath}
\usepackage{amsfonts}
\usepackage{amssymb}
\usepackage{mathtools}
\usepackage{listings}
\usepackage{mathrsfs}
\usepackage{times}
\usepackage{float}
\usepackage{color}
\usepackage{setspace}
\usepackage{color,soul}

\usepackage{amsmath,amsfonts,amsthm,eucal}
\usepackage{enumerate}
\usepackage[letterpaper,margin=3cm]{geometry}
\usepackage{float}
\usepackage[title]{appendix}
\usepackage{color}
\usepackage{tabularx}
\usepackage{siunitx}
\usepackage[tableposition=below]{caption}
\usepackage{array}

\usepackage[
pdfstartview=XYZ,
bookmarks=true,
colorlinks=true,
linkcolor=blue,
urlcolor=blue,
citecolor=blue,
pdftex,
bookmarks=true,
linktocpage=true,   % makes the page number as hyperlink in table of content
hyperindex=true
]{hyperref}

\usepackage{lipsum}
\usepackage{lineno}

\usepackage[FIGBOTCAP,TABTOPCAP,bf,tight]{subfigure}

\usepackage{natbib}
\setcitestyle{authoryear}

\usepackage[pdftex]{graphicx}
\usepackage{epstopdf}
\usepackage{booktabs} 
\usepackage{tikz,mathpazo}
\usetikzlibrary{shapes.geometric, arrows}
\usepackage{caption}

\newcommand{\tensor}[1]{\ensuremath{\boldsymbol{#1}}}

\usepackage{algorithm}
\usepackage{algorithmicx}
\usepackage[noend]{algpseudocode}

\theoremstyle{remark}

\theoremstyle{definition}

\usepackage{algorithm}
\usepackage[noend]{algpseudocode}
\newcolumntype{M}[1]{>{\centering\arraybackslash}m{#1}}

\title{Denoising diffusion algorithm for inverse design of microstructures with fine-tuned nonlinear material properties} 

\begin{document}

\author{Nikolaos N. Vlassis\thanks{Department of Civil Engineering and Engineering Mechanics, Columbia University, New York, NY 10027, \textit{Email: nnv2102@columbia.edu}} \and WaiChing Sun \footnotemark[1]\: \thanks{Corresponding author, \textit{Email: wsun@columbia.edu}}}

\maketitle

\begin{abstract}
In this paper, we introduce a denoising diffusion algorithm to discover microstructures with nonlinear fine-tuned properties. 
Denoising diffusion probabilistic models are generative models that use diffusion-based dynamics to gradually denoise images and generate realistic synthetic samples.
By learning the reverse of a Markov diffusion process,  we design an artificial intelligence to efficiently manipulate the topology of microstructures to generate a massive number of prototypes that exhibit constitutive responses sufficiently close to designated nonlinear constitutive responses. 
To identify the subset of microstructures with sufficiently precise fine-tuned properties,  
 a convolution neural network surrogate is trained to replace high-fidelity finite element simulations to filter out prototypes outside the admissible range. 
The results of this study indicate that the denoising diffusion process is capable of creating microstructures of fine-tuned nonlinear material properties within the latent space of the training data. More importantly, the resulting algorithm can be easily extended to incorporate additional topological and geometric modifications by introducing high-dimensional structures embedded in the latent space. The algorithm is tested on the open-source mechanical MNIST data set \citep{lejeune2020mechanical}. Consequently, this algorithm is not only capable of performing inverse design of nonlinear effective media but also learns the nonlinear structure-property map to quantitatively understand the multiscale interplay among the geometry and topology and their effective macroscopic properties.
\end{abstract}

\section{Introduction}
\label{intro}

Effective constitutive behaviors of manufactured and natural materials are governed by the material properties of the constituents, as well as the topology and geometry of microstructures \citep{sigmund2013topology, kumar2020inverse, wegst2015bioinspired}. 
Understanding this interplay between microstructures and macroscopic constitutive behaviors is crucial for optimizing microstructures to achieve desirable properties for a wide variety of engineering purposes across length scales. Conventionally, designing a microstructure with a designated set of material responses often requires one to solve optimization problems in the spatial domain where a level set or a phase field are used as indicator functions for the different constituents. 
This setup often leads to a multi-objective non-convex optimization problem which is non-trivial to solve or to locate the global optimal point. This difficulty motivates the use of machine learning techniques to either embed the design space onto a lower dimensional latent space or reduce the number of costly high-resolution simulations via surrogates \citep{chi2021universal, zhang2021tonr, senhora2022machine}.
On the other hand, \citet{woldseth2022use} argue that generative methods such as generative adversarial networks (GAN) and variational autoencoders (VAE) could face challenges for small perturbations of noise, especially when these small changes may lead to profound difference in the optimized structures. 

In this work, we introduce a fundamentally different approach in which the design of the microstructures is neither completed by solving a constrained optimization neither in the physical nor the latent space. 
Instead, the inverse design of a microstructure is completed through a reverse diffusion process learned by a denoising diffusion probabilistic model (DDPM)
in which we utilize embedded feature vectors of the target attributes to guide the diffusion generative process. 
Compared to the generative adversarial networks, the gradual denoising diffusion process makes the training 
more stable and the performance more scalable with increasing data size \citep{nichol2021improved}. DDPMs also does not suffer from the unstable training commonly found in the adversarial training of GAN.  
While these salient features could potentially be important for material design, the possibility of using the diffusion process to design microstructures has not been explored, according to the best knowledge of the authors. 
To fill this knowledge gap, 
we introduce context feature vectors to control both mechanical behaviors and topological features of the microstructures, allowing for the targeted inverse design of microstructures with desired nonlinear material properties properties.  
The relatively stable trajectory of the denoising process enables us to 
generate a large number of microstructure prototypes with desired macroscopic hyperelastic responses and topology characteristics such that the optimal microstructures 
can be searched by ranking the performance of the prototypes. 
To filter out the prototypes outside the desired property range,  we develop a convolutional neural network surrogate that predicts the macroscopic hyperelastic energy functional behavior to efficiently rank the generated structures without the need of running full-scale finite element simulations. 

To reduce the dimensions of the design space and to enable third-party validation, we use an open source mechanical MNIST data set provided in \citet{lejeune2020mechanical} as our training data set.  We then demonstrate the capacity of the proposed approach to accurately generate microstructures with similar energy responses to those present in the dataset.  
Finally, we explore the algorithm's ability to generate microstructure twins for both the energy response and topology features. Even when given the topology features outside of the training dataset, the DDPM model is capable of designing microstructures that exhibit the target macroscopic elastic responses. 

\subsection{Past literature on generative methods}
In the recent years, there have been significant advances in algorithms generating synthetic data samples in a wide range of applications,  from generating realistic images and videos to creating synthetic voices and music. 
The most popular include variational autoencoders \citep{kingma2013auto, vahdat2020nvae} --  learning to represent input data in a lower-dimensional latent space as probabilistic distribution and sampling from it to generate new samples,  autoregressive models \citep{kalchbrenner2017video,menick2018generating,razavi2019generating} -- learning to generate new data samples by sequentially predicting each element of the sample conditioned on the previously generated elements, and generative adverserial networks \citep{courville2014generative, karras2017progressive, brock2018large} -- a generator model learns to generate samples while a discriminator model decides if they are realistic.
Recently,  there has been a great influx of denoising diffusion probabilistic models \citep{sohl2015deep,song2020improved,ho2020denoising,nichol2021improved,ramesh2022hierarchical,rombach2022high} -- learning the reverse of a diffusion process modeled as a Markov chain to iteratively denoise a data structure and generate new samples,  that are replacing many of these models in state of the art.

Several works from the literature have utilized artificial intelligence methods to guide microstructure generation, including the use of variational autoencoders and Gaussian process regression,  generative adversarial networks, and reinforcement learning.
\citep{wang2020deep} utilize a variational autoencoder and a regressor for property prediction that provide a distance metric to measure shape similarity, enable interpolation between microstructures and encode patterns of variation in geometries and properties.
\citep{kim2021exploration} design a variational autoencoder to generate a continuous microstructure space, explore structure-property relationships and predict mechanical properties of dual-phase steels with high accuracy using Gaussian process regression.
Many have favored the use of GANs as the discriminator network component of the architecture is readily available to condition the generator component's predictions to be physical and have targeted properties.
\citep{chun2020deep} use GANs to generate ensembles of synthetic microstructures of heterogeneous energetic materials, which can be used to quantify hot spot ignition and growth and to engineer energetic materials for targeted performance.
\citep{kench2021generating} can synthesize high fidelity 3D datasets using a single representative 2D image, and generated samples of arbitrarily large volumes.
\citep{nguyen2022synthesizing} combine GANs and actor-critic reinforcement learning to synthesize realistic three-dimensional microstructures with controlled structural properties, enabling the generation of microstructures that resemble the appearances of real specimens.
\cite{kobeissi2022enhancing} use a style-based generative adversarial network with an adaptive discriminator augmentation mechanism that can successfully generate authentic patterns, which can be used to augment the training dataset for finite element simulations of soft-tissue materials. 
However,  GANs can be often difficult to train due to issues such as mode collapse, instability, and sensitivity to hyperparameters \citep{salimans2016improved,arjovsky2017towards}.
Furthermore,  diffusion models have shown potential in surpassing GANs as the new state-of-the-art for image synthesis on several metrics and data sets \citep{dhariwal2021diffusion}, as GANs can trade off diversity for fidelity and do not have good coverage of the entire data distribution.

\subsection{Organization of this article}
The rest of the paper is organized as follows.
In Section~\ref{sec:conditional_diffusion}, we discuss the concept of denoising diffusion probabilistic models for image generation, explore the problem definition of microstructure generation using denoising diffusion algorithms, and describe the neural network setup perform this task. 
In Section~\ref{sec:hyperelastic_nn}, we describe a CNN architecture that predicts the hyperelastic energy functional behavior under uniaxial extension, using the microstructure's image as input.
In Section~\ref{sec:behavior}, we demonstrate how the denoising diffusion algorithm can be trained to generate targeted microstructures with desired constitutive responses by conditioning the microstructure generation process with the hyperelastic energy functional curves.
In Section~\ref{sec:topology}, we introduce another context module neural network architecture that provides a feature vector for the targeted topology, which, along with the behavior context module, guides the microstructure generation. We show its capacity to generate microstructure twins with similar energy responses and topology features as well as test its ability to generate microstructures of topologies outside of the training data set.
In Section~\ref{sec:conclucion}, we provide concluding remarks.
For completeness, the open-source database we used to generating the latent design space is described in Appendix \ref{sec:MNIST_database}.

\section{Conditional diffusion for microstructure generation}
\label{sec:conditional_diffusion}
In this section, we provide a detailed account on how we extend the 
denoising diffusion probabilistic model, originally proposed for image synthesis, to generate microstructures with fine-tuned nonlinear material properties. 
We first explain the key ideas and mechanisms of the denoising diffusion probabilistic models and formulate the problem definition for the unconditional microstructure synthesis (Section \ref{sec:ddpm}).
We then formulate a learning problem with the specific task of the conditional generation mircostructures with target nonlinear material properties (Section \ref{sec:problem_definition}). 
Finally, we describe the neural network architecture used to learn the denoising generative process and highlight how we incorporate the embedded feature vectors that control the mechanical responses and the topology (Section \ref{sec:diffusion_architecture}).

\subsection{Unconditional generation of microstructures}
\label{sec:ddpm}

\begin{figure}[h!]
\centering
\includegraphics[width=.85\textwidth ,angle=0]{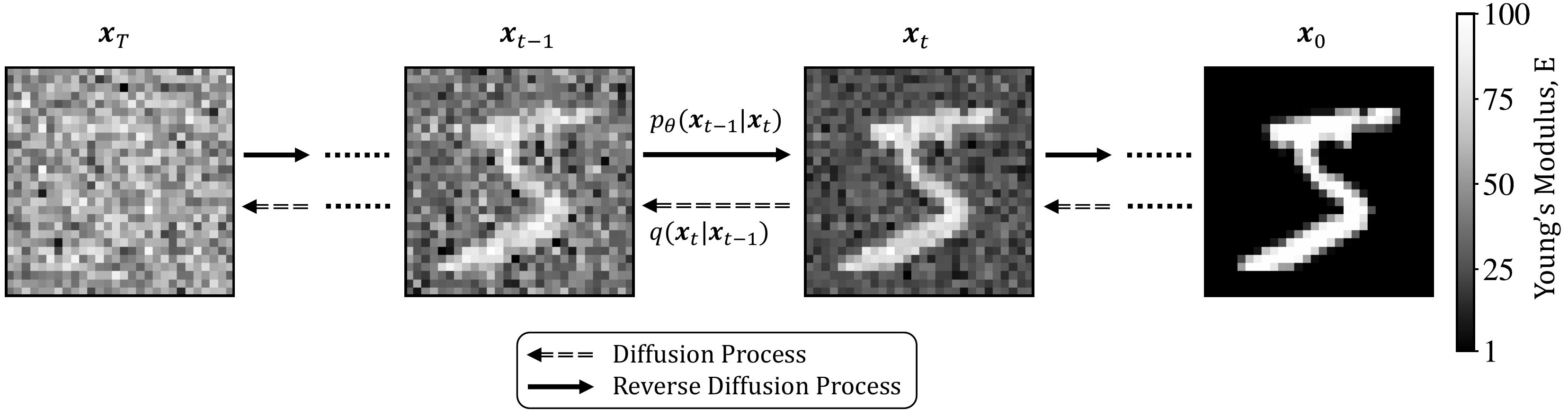} 
\caption{The diffusion process and its reverse for the elastic property distribution of a microstructure.  Noise is added to completely destroy the signal of the original distribution.  The forward noising process $q$ is fixed.  The denoising process $p_\theta$ is learned by a neural network.}
\label{fig:diffusion_process}
\end{figure}

Diffusion probabilistic models, first introduced by \citet{sohl2015deep}, are a class of generative models that aim to match a given data distribution by learning to reverse a gradual, multi-step noising process.  \citet{ho2020denoising} proposes a specific parameterization of the generative models that simplifies the training process and establishes the equivalence between denoising diffusion probabilistic models (DDPMs) and score-based generative models, which learn a gradient of the log-density of the data distribution using denoising score matching \citep{hyvarinen2005estimation}.  
Several related research works have shown that DDPMs can generate highly realistic images \citep{song2020denoising,ramesh2022hierarchical,rombach2022high}, audio \citep{chen2020wavegrad,kong2020diffwave} and forecast time series \citep{rasul2021autoregressive}.
The outstanding performance could be attributed to the gradual and iterative nature of the diffusion (see Fig. \ref{fig:diffusion_process}). 

A denoising diffusion probabilistic model for image generation can be defined as a generative model that combines the properties of denoising autoencoders and diffusion processes to generate new images. 
These models learn a Markov chain process -- where the probability of each transition depends only on the current state and time elapsed. 
They are trained in a supervised learning setting in which a variational inference is learned to produce samples that match the training data after a finite number of time steps.
The transitions of the chain are learned to reverse a diffusion process, which adds noise to the data in the opposite direction of sampling until the signal is destroyed.
In each diffusion step, the diffusion probabilistic model adds a small amount of Gaussian noise, and allows a simple neural network parameterization of the new data with the added noise. This diffusion provides a progressive trajectory from a complex data sample to Gaussian noise and thus eases the learning of the reverse denoising process by breaking it down with multiple intermediate steps as shown in Fig.~\ref{fig:diffusion_process}.
Using a continuous-time approximation of the reverse of the Markov noising process, the model smoothly traverses the latent space and generates new images.

It is highlighted that adding and removing noise in the classical image synthesis context refers to manipulating the color channels of the original image.
In the context of microstructure generation for this work, a channel represents the distribution of a material property in a specified material parameter range.
For example, for the data set used in this work the grayscale images have one channel that represents the distribution of the Young's modulus (Fig.~\ref{fig:diffusion_process}) as described in Appendix~\ref{sec:MNIST_database}.
The dimensionality of the problem can be increased by expanding the noise dimensions to represent more material parameters to facilitate the concurrent solution for multiple design targets and will be considered in future work.

In this work, we incorporate the denoising diffusion probabilistic model first introduced in \cite{nichol2021improved} to guide the generation of microstructures.   
For completeness, we provide an overview of the background for the unconditional diffusion process as described in \cite{ho2020denoising} and \cite{nichol2021improved}.
The unconditional diffusion process can be used to generate realistic microstructures in the general data range. For instance, it can be used to generate a new microstructures that are
considered similar to the microstructures provided to train the AI \citep{li2018transfer, wang2020mining, nguyen2022synthesizing}. 

Unconditional generation of microstructures alone can be useful in populating the data set with synthetic data for other machine learning tasks, such as establishing a response surface when real data are scarce. In this wok, our focus, nevertheless is on the conditional microstructure generation tasks, where the goal is not to create just any microstructures but ones that hold a particular set of topological and constitutive features demanded by the users of the DDPM. 
In practice, the conditional and unconditional generation strategy is not necessarily employed in a mutually exclusive manner. In Sections~\ref{sec:behavior} \& \ref{sec:topology}, 
we will formulate a technique in which the guided diffusion algorithm may drop the guidance context vectors (the vectors that provide additional constraints) by a percentage to enrich the generative capacity of our algorithm.
For a more complete implementation of the algorithm and the detailed differences between formulations, readers are referred to \citet{sohl2015deep, ho2020denoising, nichol2021improved}.

Inspired by the non-equilibrium thermodynamics (cf. \citet{sohl2015deep}), a diffusion model consists of two steps. The forward noising/diffusion process that slowly destroy structures of a data distribution and a reverse process that gradually restores the structures.  

\subsubsection{Forward process}
Consider a data distribution $x_0 \sim q(x_0)$, where $x_0$ is the original material property distribution sample.
In the forward noising process $q$, we produce noised samples $x_1,...,x_T$ for $T$ diffusion time steps by adding Gaussian noise in each time step. The noising is a Markov process where every noised sample $x_t$ is given by:
\begin{equation}
q(x_t | x_{t-1}) := \mathcal{N}(x_t; \sqrt{1-\beta_t} x_{t-1}, \beta_t \tensor{I}),
\label{eq:noising}
\end{equation}
where $\beta_t \in (0,1)$ is a variance schedule. 
By setting $\alpha_{t} := 1-\beta_t$ and $\bar{\alpha}_t:=\prod_{s=0}^{t}\alpha_s$, it is shown that an arbitrary step of the distribution in Eq.~\eqref{eq:noising} can be written as:
\begin{equation}
q(x_t|x_0) = \mathcal{N}(x_t; \sqrt{\bar{\alpha}_t} x_0, (1-\bar{\alpha}_t) \tensor{I}),\label{eq:noising_at}
\end{equation}
and a sample can be given by:
\begin{equation}
x_t =  \sqrt{\bar{\alpha}_t} x_0 + \epsilon  \sqrt{1-\bar{\alpha}_t},
\label{eq:sample_at}
\end{equation}
where $\epsilon \sim (0, \tensor{I})$. 
Using Bayes theorem,   \cite{ho2020denoising} find the posterior 
\begin{equation}
q(x_{t-1}|x_t,x_0) = \mathcal{N}(x_{t-1}; \tilde{\mu}(x_t, x_0), \tilde{\beta}_t \tensor{I}), 
\label{eq:posterior}
\end{equation}
where the mean of the Gaussian is:
\begin{equation}
\tilde{\mu}_t(x_t,x_0) := \frac{\sqrt{\bar{\alpha}_{t-1}}\beta_t}{1-\bar{\alpha}_t}x_0 + \frac{\sqrt{\alpha_t}(1-\bar{\alpha}_{t-1})}{1-\bar{\alpha}_t} x_t,
\label{eq:mu}
\end{equation}
and the variance is:
\begin{equation}
\tilde{\beta}_t := \frac{1-\bar{\alpha}_{t-1}}{1-\bar{\alpha}_t} \beta_t.
\label{eq:var}
\end{equation}

\subsubsection{Reverse diffusion process}
If we can learn the reverse distribution $q(x_{t-1}|x_t)$, we can sample backwards from step $t = T$ to $t=0$ to gradually remove the noise and generate a microstructure sample. 
\cite{sohl2015deep} show that this is feasible since when $T\to \infty$ and $\beta_t \to 0$,  $q(x_{t-1}|x_t)$ approaches a diagonal Gaussian distribution and a network can learn the mean $\mu_\theta$ and the diagonal covariance $\Sigma_\theta$ such that:
\begin{equation}
p_{\theta}(x_{t-1}|x_t) := \mathcal{N}(x_{t-1}; \mu_{\theta}(x_t, t), \Sigma_{\theta}(x_t, t)).
\label{eq:p_theta}
\end{equation}

While \cite{ho2020denoising} set to only learn $\mu_\theta$ and select a constant $\Sigma_\theta$, we follow the improved formulation by \cite{nichol2021improved} to parametrize both as this was observed to achieve better log-likelihoods. 
Specifically, in this formulation, we learn $\epsilon_\theta(x_t,t)$ such that:
\begin{equation}
\mu_{\theta}(x_t, t) = \frac{1}{\sqrt{\alpha_t}} \left( x_t - \frac{\beta_t}{\sqrt{1-\bar{\alpha}_t}} \epsilon_{\theta}(x_t, t) \right),
\label{eq:mu_theta}
\end{equation}
and additional to sample $x_t$ the model outputs a vector $v$ to learn $\Sigma_\theta(x_t,t)$ such that:
\begin{equation}
\Sigma_{\theta}(x_t, t) = \exp(v \log \beta_t + (1-v) \log \tilde{\beta}_t).
\label{eq:sigma}
\end{equation}

Thus, $q$ and $p_\theta$ constitute a variational autoencoder similar to \cite{kingma2013auto} to be trained by optimizing a variational lower bound for $p_\theta$.
The training of the model optimizes two objectives.
For the first learning objective, we randomly sample $t$ and optimize $\mu_\theta$ using the expectation:
 \begin{equation}
L_{\mu} = E_{t,x_0,\epsilon}\left[ || \epsilon - \epsilon_{\theta}(x_t, t) ||^2 \right],
\label{eq:L_mu}
\end{equation}
to estimate a variational lower bound.
The second learning objective to only optimize the learned $\Sigma_{\theta}$ is:
 \begin{equation}
 L_{\text{vlb}} := L_0 + L_1 + ... + L_{T-1} + L_T, 
 \label{eq:Lvlb}
\end{equation}
where:
\begin{equation}
L_t := 
\begin{cases}
    -\log p_{\theta}(x_0 | x_1) & \text{if } t=0 \\
    D_{KL}{q(x_{t-1}|x_t,x_0)}{p_{\theta}(x_{t-1}|x_t)}& \text{if } 0 < t < T \\
    D_{KL}{q(x_T | x_0)}{p(x_T)} & \text{if } t =T
\end{cases}.
\label{Ls}
\end{equation}
In the above, each term is a $KL$ divergence between the Gaussian distributions of two successive denoising steps.  The learning objective to learn both $\mu_\theta$ and $\Sigma_\theta$ is defined as:
\begin{equation}
L_{\text{hybrid}} = L_{\mu} + \lambda L_{\text{vlb}},
\end{equation}
where $\lambda = 0.001$ is a scaling factor selected to weigh the two objectives similar to  \cite{nichol2021improved}.

\subsection{Conditional diffusion with embedded feature vectors}
\label{sec:problem_definition}

While the unconditional diffusion described in the previous section can readily generate microstructures consistent with the training data set, our goal is to design microstructures 
that exhibit prescribed mechanical behaviors. To achieve this goal,  we usa a conditional diffusion process which fine-tunes the resultant microstructures via feature vectors. 

We enforce the designated material properties by introducing context into the synthesis problem,  following a relevant approach that utilized image classes or textual descriptions to provide context to target specific images \citep{ramesh2022hierarchical}. 
Our approach, however, differs in that we aim to introduce context to control both targeted mechanical behaviors and targeted topologies, and we do so by incorporating context in the form of feature vectors into the denoising diffusion algorithm. Along the time embedding feature vector that is common in diffusion algorithms,  our algorithm additionally employs a mechanical feature vector for every desired target property. This treatment allows us to modularize the problem and control different properties of the microstructures in a separate and distinct manner.
For every desired target property, we introduce a feature vector and a corresponding neural network context module that will be trained in tandem with the diffusion algorithm.

\begin{figure}[h!]
\centering
\includegraphics[width=.85\textwidth ,angle=0]{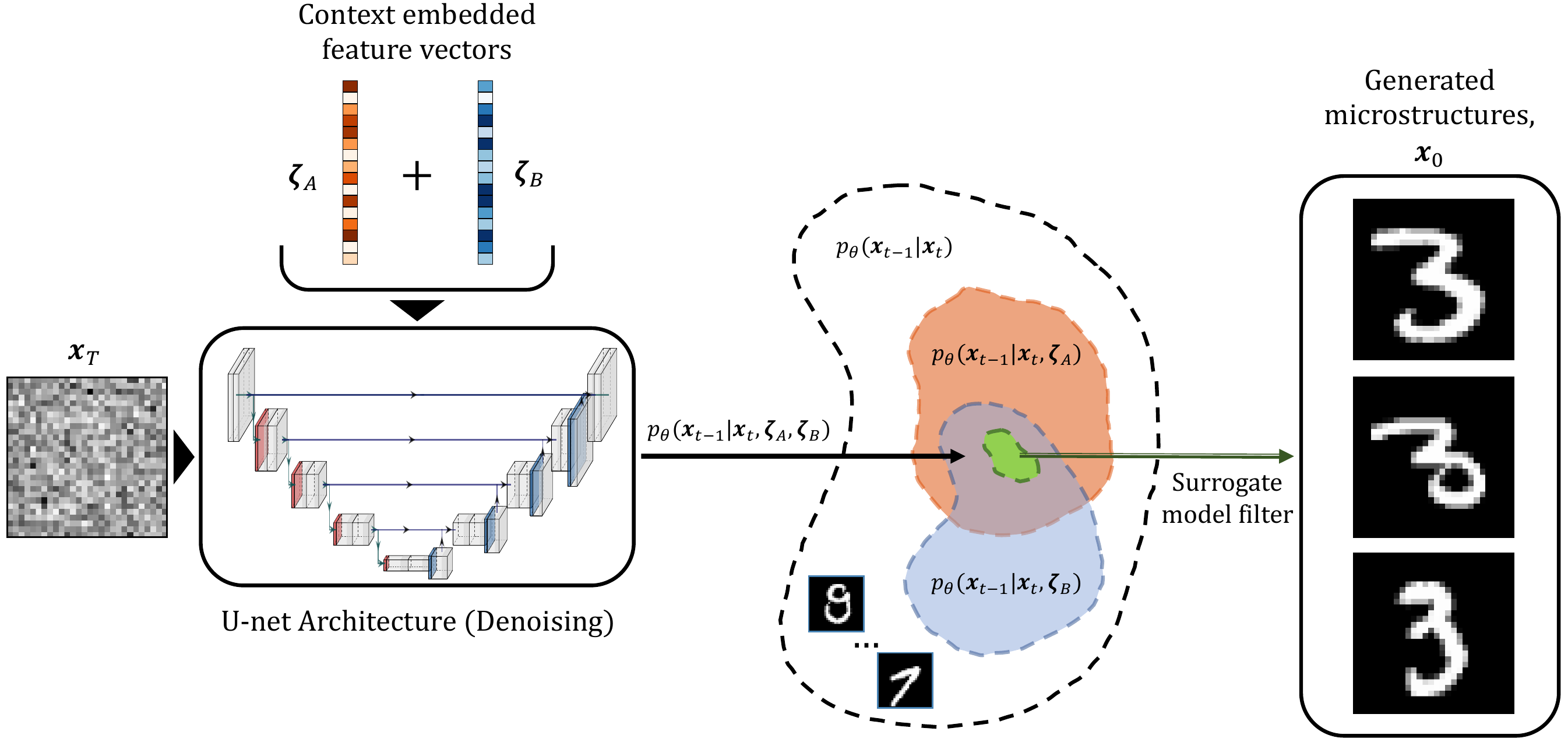} 
\caption{Schematic of the sampling framework for microstructure generation conditioned by context feature vectors $\tensor{\zeta}_A$ and $\tensor{\zeta}_B$.
The generated microstructures are further filtered by a surrogate model to demonstrate macroscopic behaviors in a desired admissible range. }
\label{fig:schematic}
\end{figure}

To incorporate the feature vector conditioning, we follow \cite{dhariwal2021diffusion} by introducing a conditional Markov process $\hat{q}$ that adds noise similar to $q$ described in the previous section.
We define the conditional noising process to:
\begin{equation}
\begin{split}
\hat{q}(x_0) := q(x_0), \quad \hat{q}(x_{t+1}|x_t,\tensor{\zeta}_1, ..., \tensor{\zeta}_N) := q(x_{t+1} | x_t),  \\
\text{ and } \hat{q}(x_1, ..., x_T | x_0,\tensor{\zeta}_1, ..., \tensor{\zeta}_N) := \prod_{t=1}^{T} \hat{q}(x_t | x_{t-1},\tensor{\zeta}_1, ..., \tensor{\zeta}_N),
\end{split}
 \label{eq:conditional_noise} 
\end{equation}
where $\{\tensor{\zeta}_1, ..., \tensor{\zeta}_N\}$ are the corresponding context feature vector embeddings for $i=1,...,N$ target microstructure properties -- be it behavior or topology. 
\cite{dhariwal2021diffusion} refer to a class conditioning given $y$ supervised labels (image classes) but we perform the conditioning with unsupervised labels that are embedded during training,  more similar to the \cite{ramesh2022hierarchical} approach. 
Thus, we are also given $\hat{q}(\tensor{\zeta}_1, ..., \tensor{\zeta}_N| x_0)$ to be the embedding generated during the training process.  It is noted that these context vectors are constant and independent of the diffusion process time. 
\cite{dhariwal2021diffusion} show that the conditional noising process with a diffusion time-independent class behaves similar to the unconditional noising process as described in the previous section and can be used to guide image synthesis.

Thus,  in order to perform the conditional sampling, we will need to train two neural network approximations. The first is the approximation $p_{\theta}(x_{t-1}|x_t)$ of $q(x_{t-1}|x_t)$ to perform the denoising as described in the previous section. The second is $p_{\phi}(\tensor{\zeta}_1, ..., \tensor{\zeta}_N| x_t)$ of $\hat{q}(\tensor{\zeta}_1, ..., \tensor{\zeta}_N| x_t)$ for the generation of the unsupervised labels.
The two will be optimized simultaneously using the training objectives in Eq.  \eqref{eq:L_mu} and Eq. \eqref{eq:Lvlb}.

In Fig.~\ref{fig:schematic}, we demonstrate a schematic of the sampling process for the conditional microstructure generation for two sample context feature vectors $\tensor{\zeta}_A$ and $\tensor{\zeta}_B$.  Each context vector corresponds to a different type of conditioning (e.g. macroscopic response and topology respectively).
At the core of the algorithm, there is a neural network architecture that learns the reverse of the diffusion process.
Starting by a random Gaussian noise input, it can iteratively denoise the sample from time $t=T$ to $t=0$ to generate a synthetic microstructure.
Without the conditioning by a feature vector, the neural network will be able to sample microstructures from entire data set latent space representation $p_{\theta}(x_{t-1}|x_t)$.
By introducing the context feature vectors, we can fine-tune the latent space the model samples from to $p_{\theta}(x_{t-1}|x_t,\tensor{\zeta}_A,\tensor{\zeta}_B)$.
To further fine-tune the microstructure generation to the desired material nonlinear properties, we introduce an additional step to filter the mass generated microstructures.
We train an surrogate model that predicts the macroscopic response of the generated microstructures, ranks the samples, and filters out the ones that do not satisfy a desired admissible range.
More details on this process are discussed in Section~\ref{sec:hyperelastic_nn}.

\subsection{Denoising diffusion neural network model architecture}
\label{sec:diffusion_architecture}

In this section, we explain (1) the neural network architecture used to perform the denoising diffusion and (2) how the context feature vectors to perform the conditioning are introduced into the diffusion process.
The architecture we used is modified from an implementation of \citet{nichol2021improved}.
They have adopted and improved on the U-net architecture used by \cite{ho2020denoising},  which have adapted the PixelCNN++ \citep{salimans2017pixelcnn++} and Wide ResNet \citep{zagoruyko2016wide} for their image synthesis tasks.
This architecture is used to approximate the denoising process as described in the previous section.

We acknowledge that the dimensionality and resolution of the images that represent the microstructure in this work -- the MNIST data set \citep{deng2012mnist}, are smaller than those studied in  \cite{nichol2021improved}, such as the CIFAR-10 \citep{krizhevsky2009learning} and ImageNet 64$\times$64 \citep{deng2009imagenet}.
Thus, we adapt a smaller architecture version of the original U-net that was deemed adequate to perform our synthesis tasks.
Specifically,  in our U-net implementation, there are two stacks that perform the downsampling and upsampling in four steps respectively.
Each step has one residual block.
We use one attention head with 32$\times$32, 16$\times$16, and 8$\times$8 attention resolutions.
We modify the output of the architecture to have one output image channel to produce the microstructure grayscale material property maps.
The U-net utilizes $[C,2C,4C,8C]$ channel widths from higher to lower resolutions with the base model channel size selected as $C=32$.  
The architecture is conditioned by the diffusion time step $t$ through an embedded feature vector $\tensor{\zeta}_t$.
This vector is produced by a sequential two-layer network -- it has two-dense layers of 128 neurons each with a Sigmoid Linear Unit (SiLU) and a Linear activation function respectively.

At this point of the architecture, we inject our context feature vectors to perform the microstructure conditioning.
Specifically, given context data structures $\tensor{Z}_i$ for $i=1,...,N$ target properties
to condition the microstructure generation with, we define neural network architectures $c_i$ that produce context feature vector embeddings $\tensor{\zeta}_i = c_i(\tensor{Z}_i)$. In this framework,  the context data structures are curves (hyperelastic energy functional responses) and 2D images (topologies).
However,  any other type of data structure can be embedded to control the generation process.
The context feature vector embeddings are introduced in the architecture by defining an embedding vector $\tensor{\zeta}_\text{emb}$ for both the diffusion time and the microstructure properties as:
\begin{equation}
\tensor{\zeta}_\text{emb} = \tensor{\zeta}_t + \tensor{\zeta}_i + ... +\tensor{\zeta}_N \text{ for } i=1,...,N. 
\label{eq:feature_vector}
\end{equation} 

In this work, we will introduce two types of context embeddings, by (1) material behavior and (2) topology in Section~\ref{sec:behavior} and Section~\ref{sec:topology} respectively.
The corresponding neural network architectures $c_i$ are defined in these sections.
We highlight that we introduce the conditioning as the summation of embedded feature vectors to allow a modular structure of the framework independent of the diffusion algorithm.
More types of conditioning will be considered in future work.

\section{Neural network for material behavior prediction}
\label{sec:hyperelastic_nn}

In order to further optimize the microstructure generation for desired material nonlinear properties,  we train a surrogate model which predicts the macroscopic response of the generated microstructures and, subsequently, ranks and filters out samples which do not fall within the desired admissible hyperelastic energy functional range.
In this section,  we describe the Convolutional Neural Network (CNN) architecture to perform this task.
Given the distribution of a microstructure's Young's modulus, the model can predict the hyperelastic energy functional behavior under uniaxial extension.
This CNN model can be used to predict the behavior of a new microstructure during a uniaxial extension simulation,  eliminating the need to perform a complete finite element simulation. 
These predictions can then be used in place of the output of an FEM simulation,  allowing for faster prediction of the microstructure's behavior with a reasonable trade-off in accuracy. 
It is noted that full-scale finite element simulations will also eventually be used to validate the generated microstructures in following sections. 
However, the CNN counterpart will be used to quickly check the estimated behavior of a microstructure. 

\begin{figure}[h!]
\centering
\includegraphics[width=.75\textwidth ,angle=0]{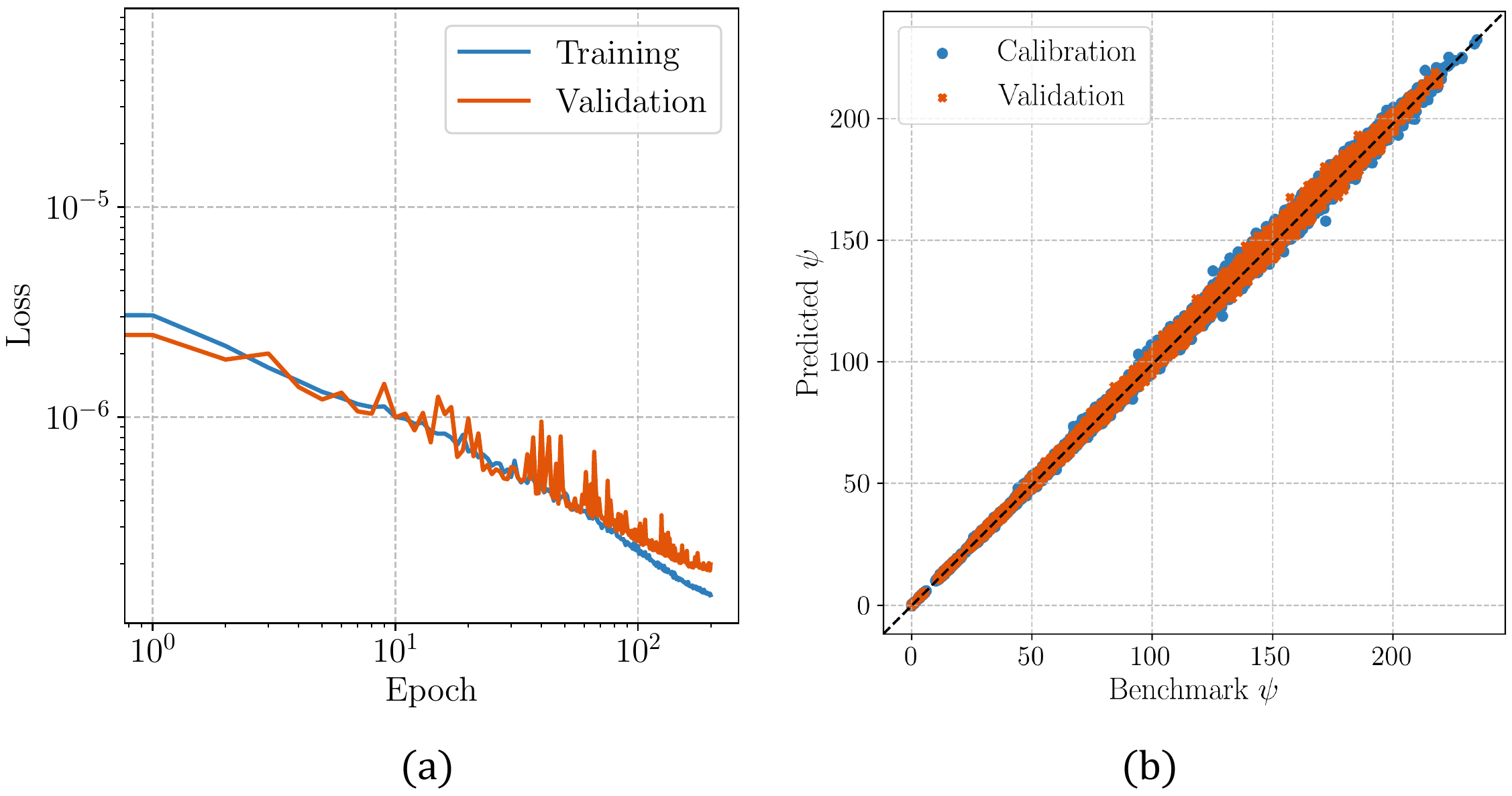} 
\caption{(a) Training and validation loss function values and (b) predictions of the energy functional response of all the training and testing samples of the MNIST dataset for the CNN architecture.}
\label{fig:forward_NN}
\end{figure}

The neural network implemented follows a standard CNN architecture and consists of two Convolutional layers with a kernel size of 3,  stride of 1, and padding of 1.  
The first CNN layer has 1 input channel and 16 output channels,  and the second Convolutional layer has 16 input channels and 16 output channels.  
The output of the first Convolutional layer is passed through a Max Pooling layer with a kernel size of 2 and stride of 2, which down-samples the input by taking the maximum value over a 2x2 window. 
The output of the second Convolutional layer is also passed through a Max Pooling layer with the same configuration. The output of the second Max Pooling layer is then flattened to a single dimension, representing the features extracted by the CNN. 
This flattened output is then passed through three fully-connected (Dense) layers with 120, 84, and 13 neurons, respectively.  
The output size of 13 corresponds to the number of time steps in the Mechanical MNIST uniaxial extension curves.
The ReLU activation function is applied to the output of the CNN and Dense hidden layers, while the output uses a Linear activation function.
The kernel weight matrices of the layers are initialized with the default Glorot uniform distribution, and the bias vectors are initialized with a zero distribution. 
The model was trained for 200 epochs with a batch size of 256 using the Adam optimization algorithm with a learning rate of $10^-3$ and a mean squared error (MSE) loss function. 
The learning rate was reduced by a factor of 0.9 when the validation loss plateaued for 5 epochs.
The network was trained on the 60,000 and validated on the 10,000 MNIST training and test samples and their corresponding energy functional $\psi$ curves of the database described in Appendix~\ref{sec:MNIST_database}.
The results of the training are demonstrated in Fig.~\ref{fig:forward_NN} where the network is shown to perform well in the training set with a good capacity to generalize outside the samples used for the calibration.  

\section{Generation of hyperelastic microstructures conditioned by material behavior}
\label{sec:behavior}

In this section, we will demonstrate that the denoising diffusion algorithm described in a Section~\ref{sec:conditional_diffusion} can be trained on the Mechanical MNIST data set to generate targeted microstructures with desired constitutive responses by conditioning the microstructure generation process with the hyperelastic energy functional curves. 
In Section~\ref{sec:curve_conditioned_training}, we will first describe the hyperelastic curve context module of the diffusion algorithm,  the training of the architecture, and the capacity of the algorithm to 
generate synthetic microstructures with similar statistical properties as the given data set.
In Section~\ref{sec:polynomial},  we will input random polynomials to design targeted microstructures outside of the training range and explore the capacity of the algorithm to generate microstructures for unseen behaviors.

\subsection{Training of behavior conditioned architecture}
\label{sec:curve_conditioned_training}

In this section,  we introduce a context module neural network architecture that extracts the behavior feature vector of the hyperelastic energy functional curves from the Mechanical MNIST dataset to guide the microstructure generation process.  We detail the training procedure and demonstrate the capacity of the architecture to accurately generate microstructures with similar energy responses to these present in the Mechanical MNIST dataset.

\begin{figure}[h!]
\centering
\includegraphics[width=.95\textwidth ,angle=0]{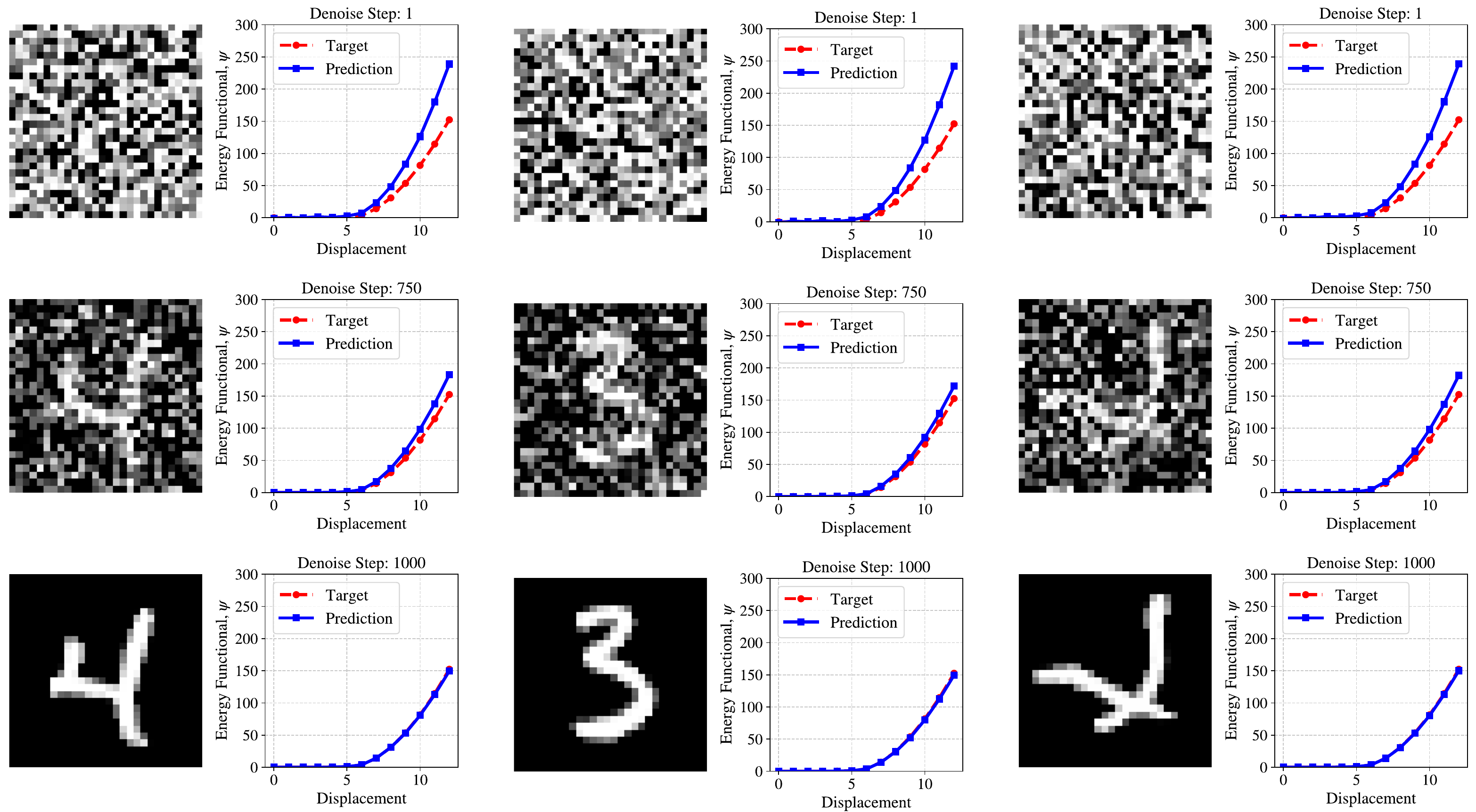} 
\caption{Microstructure generation and neural network prediction results for the same hyperelastic energy functional response from the Mechanical MNIST testing dataset. The results are shown for denoising steps 1, 750,  and 1000.}
\label{fig:same_curve}
\end{figure}

The context module in this section embeds the hyperelastic energy functional curves described in Appendix~\ref{sec:MNIST_database} into feature vectors.
The module follows a simple feed-forward neural network architecture.
It inputs the energy functional values $\psi$ of the 13 time steps of the uniaxial extension experiments of the Mechanical MNIST data set.
It is noted that here we are only using the energy functional values since the displacement inputs for all the FEM simulations are the same and this input information would be redundant as a context for this case of microstructure generation.
In the case the input strain was variable or there existed different types of tests in the database, we would have to input additional identifiers to properly control the microstructure generation.
The sequential module has two Dense layers of 128 neurons each.  The hidden layer has a SiLU activation function and the output feature vector layer has a Linear activation function.
We opt for an embedded feature vector of 128 neurons that matches the dimensions of the time embedding vector.
The predicted behavior feature vector is added to the time embedding feature vector as described in Section~\ref{sec:diffusion_architecture}.
The kernel weight matrices of the layers are initialized with the default Glorot uniform distribution, and the bias vectors are initialized with a zero distribution.

After assembling the diffusion architecture with the behavior context module, we train the diffusion model for 50,000 steps on the 60,000 sample pairs of MNIST handwritten digits and energy functional curves. We also set the context to be dropped with a 10\% chance -- the feature vector used is $\tensor{\zeta_\text{emb}=\tensor{\zeta}_t}$-- to improve the algorithm's capacity to perform unconditional synthesis. 
We opt for 1000 diffusion time steps with a linear noise schedule as used in \cite{ho2020denoising}.  We use an Adam optimizer \citep{kingma2014adam} with a learning rate of $10^{-4}$ with a batch size of 128.

\begin{figure}[h!]
\centering
\includegraphics[width=.75\textwidth ,angle=0]{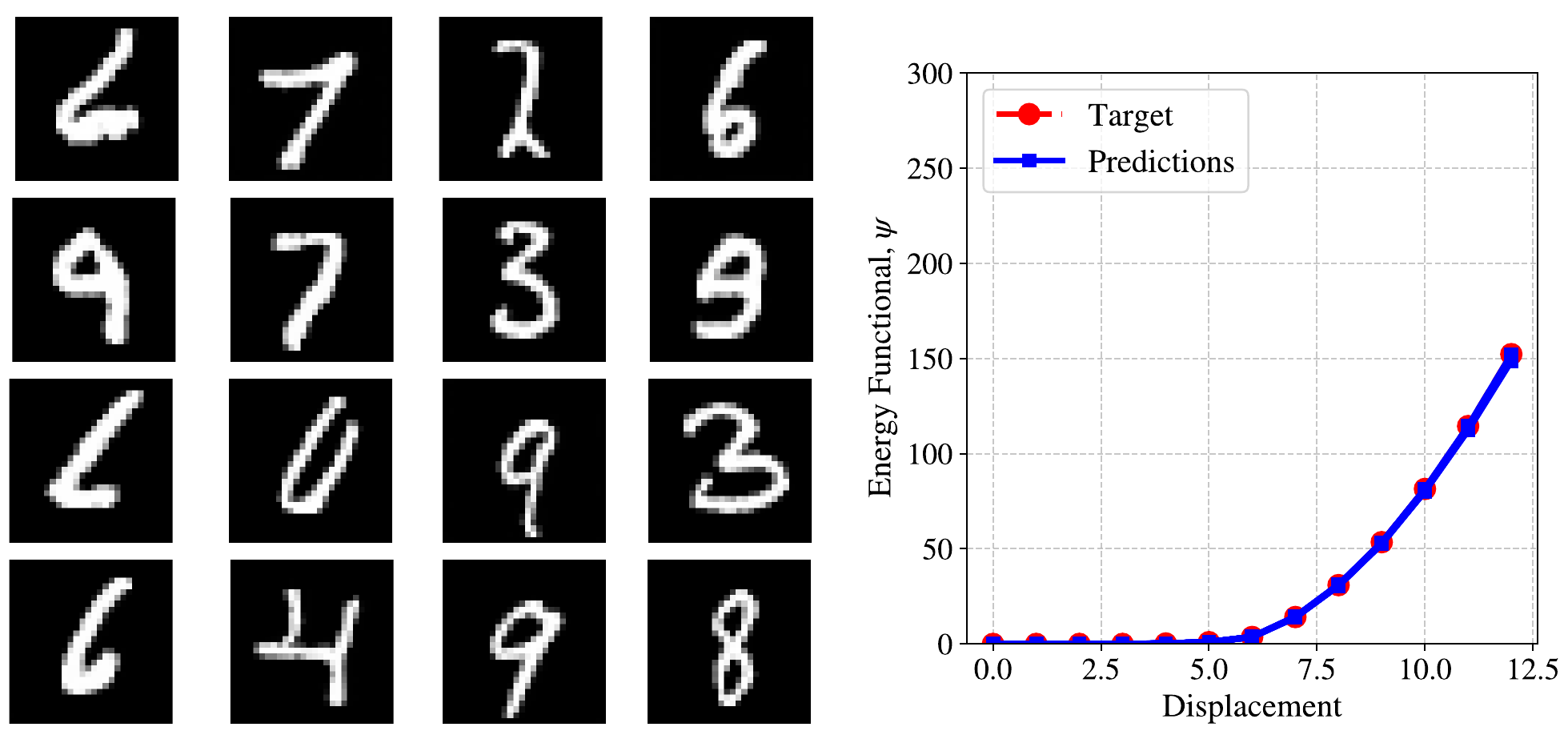} 
\caption{Microstructure generation and neural network prediction results of 16 samples for the same hyperelastic energy functional response from the mechanical MNIST testing data set.  The samples were batch generated and filtered to have an energy functional prediction with MSE$<10^{-4}$.}
\label{fig:same_curve_16}
\end{figure}

In Fig.~\ref{fig:same_curve}, we observe that the algorithm can create twins for a specific behavior that do not look alike or do not resemble the topologies in the training data set.
All three microstructures are generated for the same mechanical response from the mechanical MNIST testing data set.
The algorithm is trained on images of handwritten digits and does not have any limitations or context guidance on the generated images to fall within a specific image class. 
This is an important aspect of the behavior guided diffusion model. 
It means that the algorithm is not constrained to generate microstructures that resemble the images it was trained on but rather has the ability to generate a wide variety of microstructures that may have different shapes, forms, and other properties. This feature allows for a greater degree of flexibility in the design of microstructures and has the potential to enable the generation of microstructures with specific properties that may not be found in the training dataset. Additionally, it also allows to generate microstructures with shapes or forms that are not seen during the training phase. This ability to generalize and adapt to new behavior context inputs is a key advantage of this approach to microstructure generation, as it enables the discovery of a diverse range that can meet specific design criteria.

It is highlighted that it is expected the generated microstructures do not necessarily correspond to a perfectly accurate energy response compared to the target one. 
This is because the generation process is probabilistic and performed using the denoising diffusion probabilistic algorithm, which inherently introduces a degree of uncertainty. 
As a result, the generated microstructures have an error range in their predictions of the energy response. 
To control the degree of accuracy of the generated microstructures, we filter out the ones that do not satisfy a target MSE error limit of MSE$<10^{-4}$ using the CNN architecture previously described in Section~\ref{sec:hyperelastic_nn}.  Thus, we can batch generate multiple microstructures and quickly select the desired ones without having to run the much slower finite element simulations to validate.
In Fig.~\ref{fig:same_curve_16}, we demonstrate 16 microstructure twins that were batch generated in this way for the same mechanical MNIST testing curve.
The error range of the generation process is further explored in the following section.

\subsection{Generate hyperelastic microstructures with polynomial energy functional}
\label{sec:polynomial}

In this section,  we will be further testing the genation for energy functional forms not included in the data set.
Instead of using the hyperelastic energy functionals of the Mechanical MNIST data set as a guide, we will be using a set of random energy functional polynomials as the target behaviors to generate microstructures. 
These random polynomials will be selected from the coefficient ranges present in the Mechanical MNIST data set and validate that the generated microstructures are physically realistic and representative of the range of behaviors present in the data set. 
By using these random polynomials as the target behaviors, we can generate a diverse set of microstructures with a wide range of hyperelastic properties, rather than being limited to the specific behaviors present in the original dataset. 
This allows for the exploration of the behavior of microstructures with a wider range of properties and to more fully understand the relationship between microstructure and behavior.

\begin{figure}[h!]
\centering
\includegraphics[width=.75\textwidth ,angle=0]{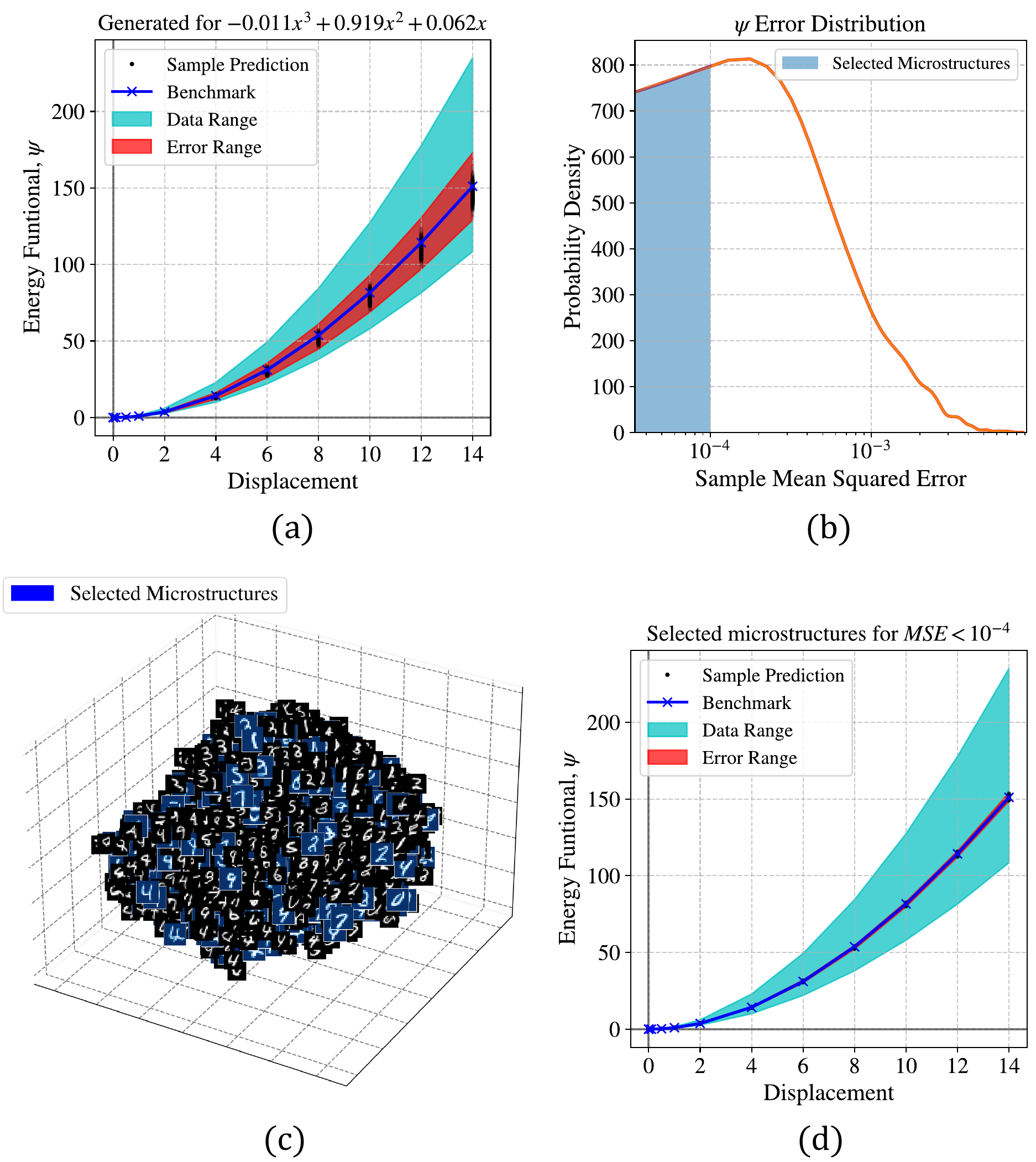} 
\caption{(a) Energy functional neural network predictions for 2112 generated microstructures for the random target polynomial $-0.011 x^3+0.919 x^2+0.062 x$.  The data range and error range of the predictions are also demonstrated.
(b) $MSE$ distribution for the predicted energy functional of the 2112 microstructures.  Microstructures with an $MSE$ less that $10^{-4}$ are selected.
(c) The 2112 generated microstructures and the selected microstructures that have an $MSE<10^{-4}$.  The microstructures are embedded using the T-distributed Stochastic Neighbor Embedding algorithm for easier visualization in three axes.
(d) Energy functional neural network predictions for the selected generated microstructures that have an $MSE<10^{-4}$. }
\label{fig:hyperelastic_polynomial_filter}
\end{figure}

\begin{figure}[h!]
\centering
\includegraphics[width=.80\textwidth ,angle=0]{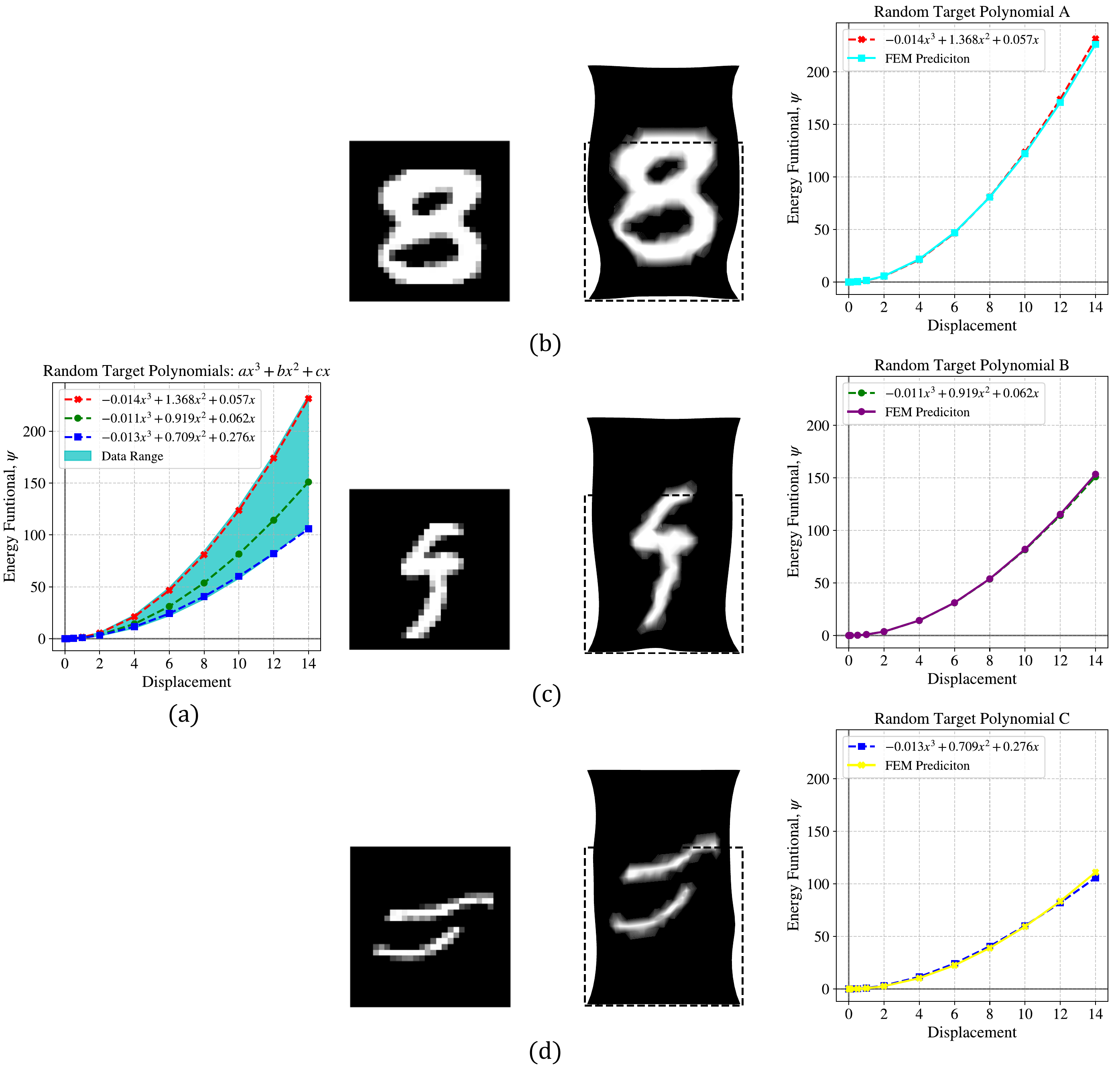} 
\caption{Microstructure generation and FEM simulation results for three random target energy functional polynomials.}
\label{fig:hyperelastic_polynomial}
\end{figure}

To generate the target polynomials for the microstructure generation process, we first fit all the energy functional curves in both the training and testing data sets with third-order polynomials.
The general form of the polynomials is $a x^3 + b x^2 + c x$. 
After fitting the curves,  we find the ranges of the coefficients $a = [-0.0162,  -0.0047]$, $b = [0.6346,  1.3968]$, and $c = [ 0.00532,  0.4114]$ 
We sample three random polynomials from these coefficient ranges to explore the entire energy data range of the data set.
From a higher to a lower energy functional range, we sample $-0.014 x^3+1.368 x^2+0.057 x$.,  $-0.011 x^3+0.919 x^2+0.062 x$, and $-0.013 x^3+0.709 x^2+0.276 x$.
The data range and the sampled energy functionals are shown in Fig.~\ref{fig:hyperelastic_polynomial}(a).

In order to generate microstructures with accurate targeted behaviors, we filter the generated microstructures using the CNN architecture previously described in Section~\ref{sec:hyperelastic_nn}.
We continuously generate microstructures until 512 of them are predicted to have an MSE$<10^{-4}$.
The error range of the predictions can be quantified and is shown in a Fig.~\ref{fig:hyperelastic_polynomial_filter}(a).
In this figure, we end up generating 2112 microstructures for the second target polynomial $-0.011 x^3+0.919 x^2+0.062 x$ as a context, and plot the error range of every predicted behavior using the CNN. We also plot the MSE distribution curve for these predictions,  shown in Fig.~\ref{fig:hyperelastic_polynomial_filter}(b). This allows to visualize the overall accuracy of the CNN in predicting the energy response of the generated microstructures, as well as the range of errors that we can expect in these predictions. 
In addition to visualizing the overall error distribution, we can also use the CNN to filter out and select only the microstructures that have a predicted MSE within a certain range. In Fig.~\ref{fig:hyperelastic_polynomial_filter}(c),  we plot the 2112 generated microstructures,  and highlight the 512 that have a predicted MSE lower than a selected $10^{-4}$ admissible limit.  We also replot the error range for this MSE limit in Fig.~\ref{fig:hyperelastic_polynomial_filter}(d), to show how the errors are distributed among the selected microstructures.  By selecting only the microstructures with a predicted MSE below a certain limit, we can effectively filter out the microstructures that have a higher degree of uncertainty, and focus on those with a higher degree of accuracy without having to run a full FEM simulation. 

The filtering procedure is performed during the microstructure generation for all the target polynomials.
The results for the generated microstructures are shown in Fig.~\ref{fig:hyperelastic_polynomial}(c), (d), and (e).
The microstructures shown were first selected to have a MSE less than $10^{-4}$. 
To validate the microstructure design, we also test the microstructure by performing a forward FEM simulation and validate that the generated microstructures indeed have the desired behaviors.
We can also observe that the shapes are progressively thicker from lower to higher polynomial value ranges. 
This is expected as the higher final energy functional values represent increasingly stiffer materials -- it is expected that more stiff inclusion material is predicted for the microstructures. 
By carefully selecting the target behavior, we can achieve a high-level control of the overall shape and topology of the generated microstructures with an accurate control of the target behavior.
In the next section, along with the target behavior we will control the topology of the generated microstructures as well.

%\FloatBarrier

\section{Generation of hyperelastic microstructures conditioned by both material behavior and topology}
\label{sec:topology}

In this section,  we introduce another context module neural network architecture that will provide another feature vector to control topology. 
Along with the behavior context module, it will be used to guide the microstructure generation for two concurrent tasks. 
We outline the training procedure and demonstrate the capacity of the architecture to generate microstructure twins now for both the energy response and topology features for the behaviors and microstructure present in the Mechanical MNIST data set.  
Additionally, we will test the capacity of the diffusion algorithm to generate microstructures of topologies outside of the training data set.

\subsection{Training of behavior and topology conditioned architecture}
\label{sec:forward_fem}

In this section, we introduce an enhanced context module neural network architecture that extracts both behavior and topology feature vectors from the Mechanical MNIST data set to guide the microstructure generation process.  In addition to the material behavior module introduced in the previous section, we now also include a topology module. We detail the training procedure and demonstrate the capacity of the enhanced architecture to accurately generate microstructures with similar energy responses and topologies to those present in the Mechanical MNIST data set.

\begin{figure}[h!]
\centering
\includegraphics[width=.80\textwidth ,angle=0]{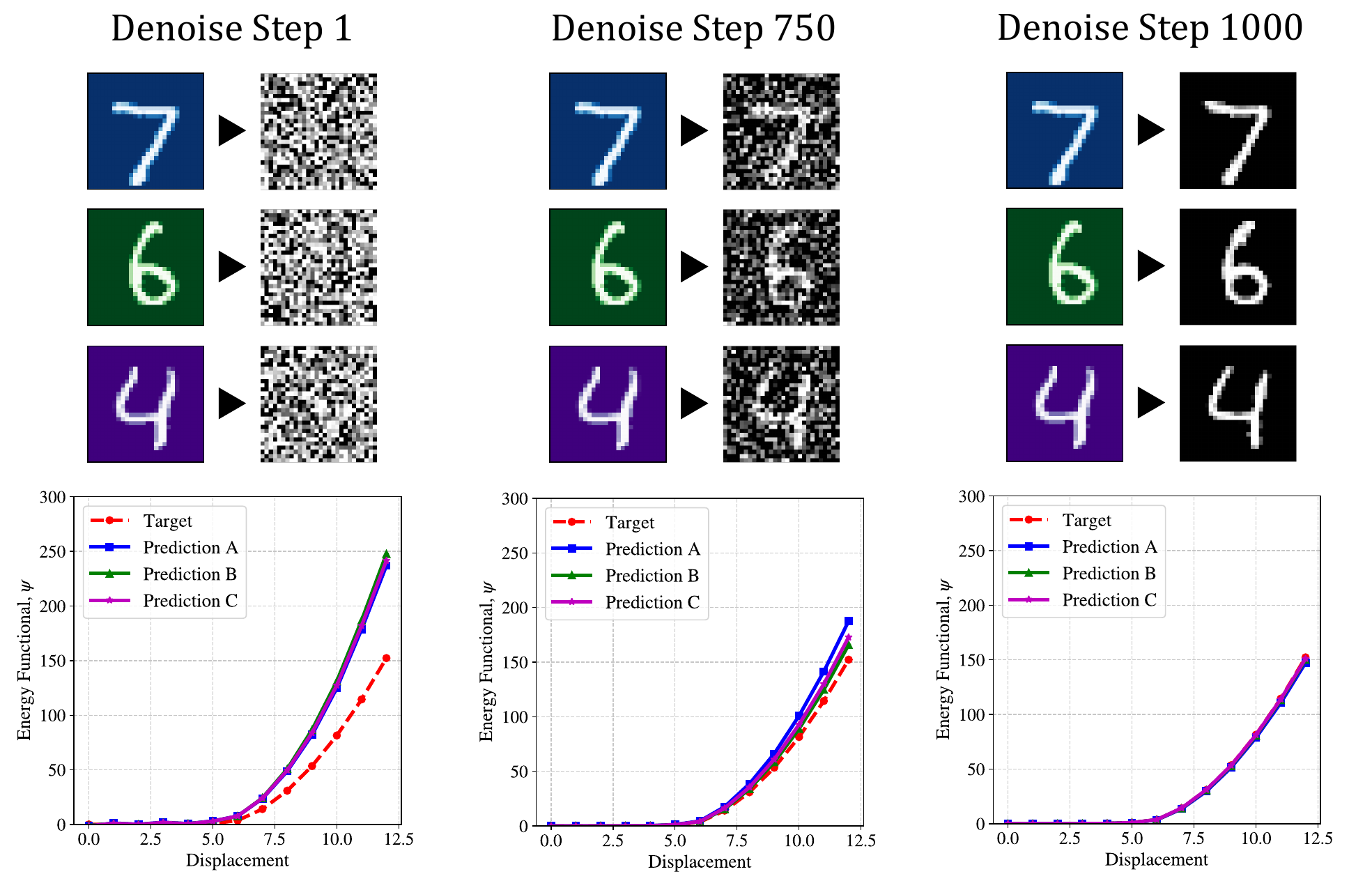} 
\caption{Generation of topology and behavior twins from a microstructure from the MNIST test set. The results are shown for steps 0, 750, and 1000 of the denoising process.}
\label{fig:mnist_test_image}
\end{figure}

The additional topology context module follows a convolutional neural network architecture.
The architecture has two convolutional layers of kernel size 3, stride 1 and padding 1 with ReLU activations.
The first layer inputs the grayscale map that describes the desired target material property distribution context.  
Both layers have 16 output channels.  Both are followed by a max pooling layer of kernel size 2 and stride 2 to down-sample their output. 
The output of the second pooling layer is then flattened and passed through three fully-connected layers with 128 neurons each with two ReLU activations and a Linear one respectively. 
This choice of 128 neurons is made as it aligns with the size of time and behavior context embedding vectors from the previous sections.
Similarly to the previous section, the context vector containing the time, behavior, and topology contexts comes from the summation of the three embedded vector counterparts.
The training procedure is identical to the one described in Section~\ref{sec:curve_conditioned_training} with the difference that along every sample we also input a topology context map.

With the addition of a topology module, along with the material behavior module, we are now able to perform more controlled microstructure generation tasks. 
In Fig.~\ref{fig:mnist_test_image}, we demonstrate the generation of microstructures from image and curve pairs from the Mechanical MNIST testing set.
In this example,  we attempt to create twins for a randomly sampled microstructure and its corresponding hyperelastic behavior.
We first check if the algorithm can reproduce a microstructure with the same topology context (the digit 7) and behavior.
We then test the capacity to generate microstructures that have the same behavior context but different topology context inputs randomly sampled from the MNIST test set (the digits 6 and 4). We can see the algorithm modifies the material property map inputs to synthesize structures to adjust to the specific target behavior.
These microstructures not only look similar to the images they were generated from but also have the same hyperelastic behavior as the target energy functional curve. 
Thus, we show that we can perform the creation of both behavior and topology twins.
While in this example the alterations from the input context images may seem slight,  we test some more apparent topology changes in the following section.

\subsection{Generation of topologies outside the training dataset}
\label{sec:forward_fem}

In this section, we will investigate the capability of the behavior and topology guided diffusion model to generate microstructures for topologies that are not represented in the Mechanical MNIST data set. 
Despite the algorithm being trained on images of handwritten digits, there are no limitations or context guidance imposed on the generated images to fall within a specific image class, as previously demonstrated in Section~\ref{sec:curve_conditioned_training}. This suggests that the algorithm has the potential to generate microstructures of other shapes and geometries as well.
However, it is observed that while the topology context input is dissimilar to the MNIST data set, the images generated in this section tend to resemble the MNIST style. 

\begin{figure}[h!]
\centering
\includegraphics[width=.95\textwidth ,angle=0]{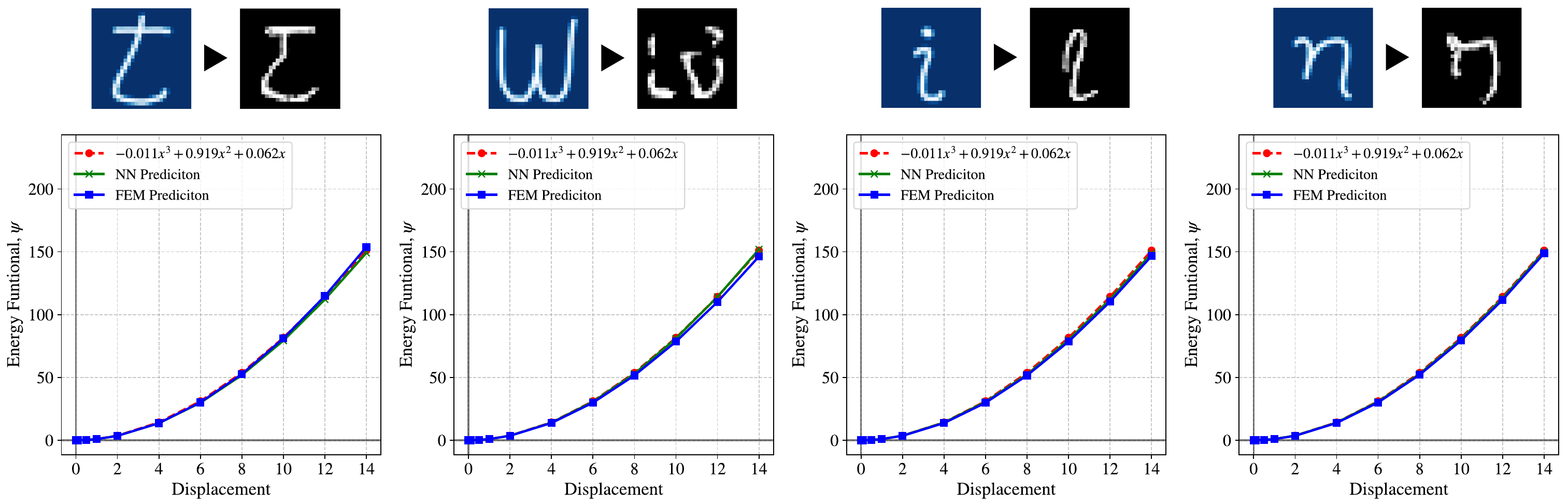} 
\caption{Microstructure generation and FEM simulation results for the target energy functional polynomial $-0.011 x^3+0.919 x^2+0.062 x$ and target images that spell the word "twin".}
\label{fig:twin}
\end{figure}

\begin{figure}[h!]
\centering
\includegraphics[width=.95\textwidth ,angle=0]{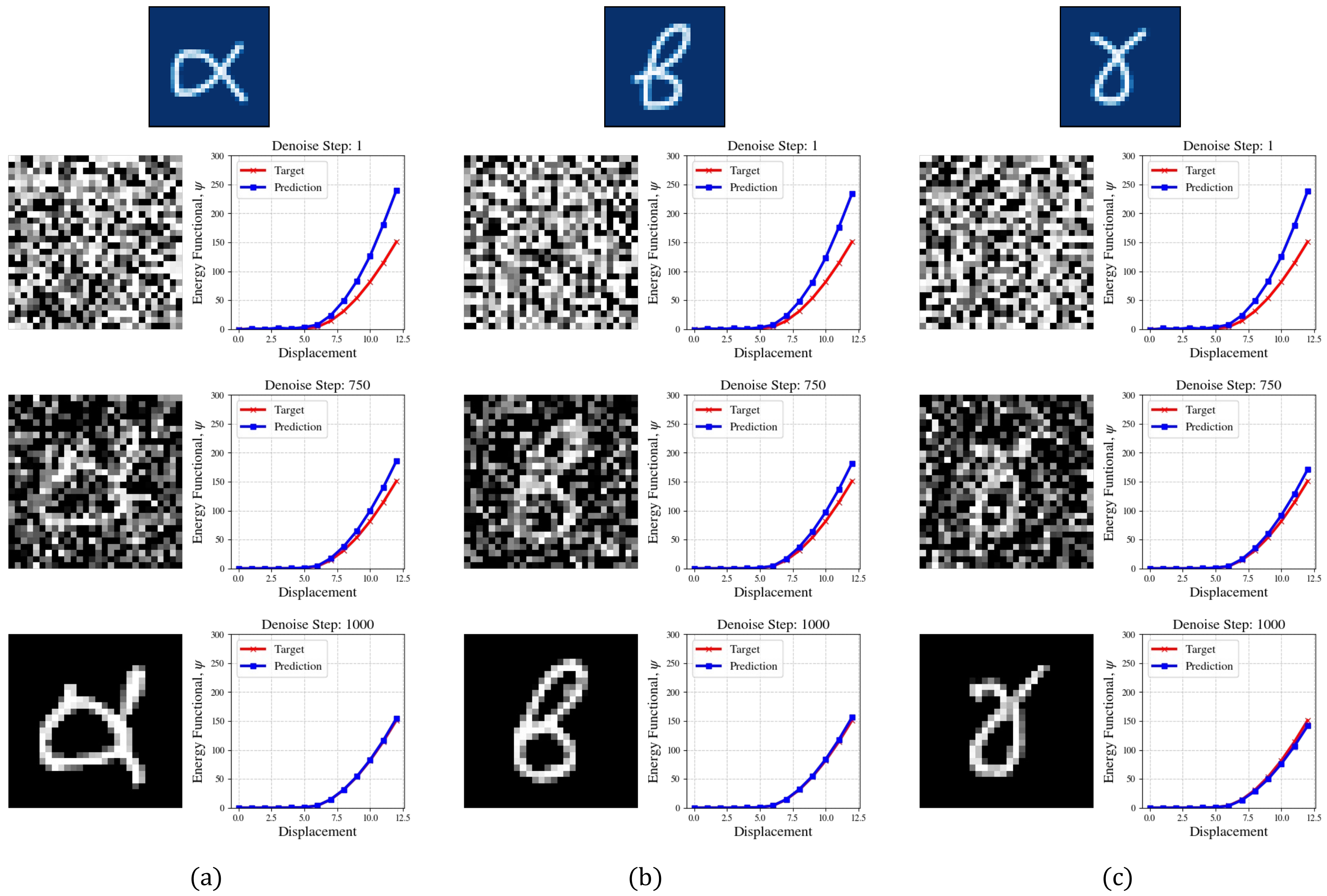} 
\caption{Microstructure generation and neural network prediction results for the target energy functional polynomial $-0.011 x^3+0.919 x^2+0.062 x$ and targeted images for the letters $\alpha$,  $\beta$, and $\gamma$. The results are shown for denoising steps 1, 750,  and 1000.}
\label{fig:alpha_beta_gamma}
\end{figure}

By focusing on the topology context aspect of image generation, we will create microstructures that share the same energy response, represented by the energy functional $-0.011 x^3+0.919 x^2+0.062 x$ -- all the microstructures in this section have the same energy response. 
Additionally, to improve the accuracy of the generated microstructures, we utilize a mass generation method of prototypes and employ a CNN model from Section~\ref{sec:hyperelastic_nn} to filter out those microstructures that have a predicted MSE$>10^{-4}$.
We were able to utilize the proposed method to generate microstructures that spell the word "twin" as illustrated in Fig.~\ref{fig:twin}.
 Furthermore, we demonstrate the denoising process, in which the letters $\alpha$,  $\beta$, and $\gamma$ were generated, as depicted in Fig.~\ref{fig:alpha_beta_gamma}. 
 These characters were not present in the training data set, however, the algorithm has a limited capability to recreate them while also conforming to the target energy functional behavior.

An analysis of the results generated by the behavior and topology guided diffusion model illustrate the potential benefits of utilizing this approach in the synthesis of microstructures. These advantages include the ability to create microstructures of specific shapes, as well as the potential to design microstructures with other properties, such as symmetry and targeted material volumes and ratios. 
Nevertheless, a current challenge in this application is the limited similarity between the topology context inputs used and the Mechanical MNIST data set. 
This has resulted in difficulty in efficiently identifying suitable microstructures, as well as instances in which it is not possible to discover microstructures with a predicted energy response with MSE $< 10^{-4}$ at all. 
Specifically, when topology context inputs are dissimilar to the MNIST data set, the error range of the generated microstructures may be larger, deeming it harder to find a suitable microstructure. 
For example, inputting a very stiff topology context image and an energy curve in the lower range of the material behaviors, it is expected that the algorithm would have difficulty in designing a microstructure that satisfies both criteria simultaneously.

One potential solution to this obstacle in discovering microstructures with dissimilar topology context inputs would be to augment the training data set. 
Specifically, by selecting a more heterogeneous data set that includes a diverse range of microstructure shapes and forms, the algorithm would have access to a wider range of reference examples upon which to base its generative process. 
Furthermore, in cases where a specific inclusion shape is desired, incorporating additional examples of that shape within the training data set could also prove beneficial in terms of improving the algorithm's ability to generate microstructures featuring that specific shape. Additionally, employing other forms of guidance within the microstructure generation process, such as through the use of metrics (e.g.  a symmetry loss metric, a target ratio of soft to stiff material) or neural networks, could also help to improve the quality of the generated microstructures. 
Such additional guidance mechanisms can provide a filter for the generated microstructures, allowing the algorithm to focus on those with the desired characteristics and eliminate those that are not similar to the target topology.

\section{Conclusion}
\label{sec:conclucion}

We present a novel approach to design microstructures with desired mechanical behaviors and topologies by introducing context feature vectors that control targeted properties in a DDPM model. 
We demonstrate the capacity of the architecture to accurately generate microstructures with similar energy responses and topologies to those present in the Mechanical MNIST data set, while demonstrating a reasonable extrapolating capacity. 
Our method is based on a modular framework where a feature vector is assigned to every desired target property. 
This allows for the incorporation of different types of target properties and their corresponding embeddings,  rendering the problem more scalable and easier to interpret. 
Future work will explore the design of microstructures with additional behavior controls such as behaviors in the microscale along the macroscopic responses and more precise topology constraints, such as symmetry. This could be achieved by introducing more complex microstructure databases as well as additional constraints in the diffusion algorithm.
The flexibility and speed of the design of microstructures using diffusion algorithms will not only facilitate the generation of digital twins but could also be extended to generate synthetic databases in usually data-scarce engineering applications.

\section{Acknowledgments}
The authors are supported by the National Science Foundation under grant contracts CMMI-1846875 and OAC-1940203, and
 the Dynamic Materials and Interactions Program from the Air Force Office of Scientific 
Research under grant contracts FA9550-19-1-0318,  FA9550-21-1-0391 and FA9550-21-1-0027, with additional 
support provided to WCS by the Department of Energy DE-NA0003962. 
These supports are gratefully acknowledged. 
The views and conclusions contained in this document are those of the authors, 
and should not be interpreted as representing the official policies, either expressed or implied, 
of the sponsors, including the Army Research Laboratory or the U.S. Government. 
The U.S. Government is authorized to reproduce and distribute reprints for 
Government purposes notwithstanding any copyright notation herein.

\section{Data and code availability}
The data and computer code that support the findings of this study are available from the corresponding author upon request. 
 
\appendix
\section{Mechanical MNIST Database}
\label{sec:MNIST_database}

\normalsize{
This work uses the Mechanical MNIST data set \citep{lejeune2020mechanical} that contains pairs of heterogeneous microstructures and their corresponding hyperelastic responses.
Mechanical MNIST is a data set of finite element simulation results generated from images in the MNIST database of handwritten digits \citep{deng2012mnist}.
The MNIST dataset is a large database of handwritten digits that is commonly used for training, evaluating, and benchmarking image classification algorithms. 
It consists of 60,000 training images and 10,000 test images, each of which is a 28x28 pixel grayscale image of a handwritten digit from 0 to 9.

\begin{figure}[h!]
\centering
\includegraphics[width=.85\textwidth ,angle=0]{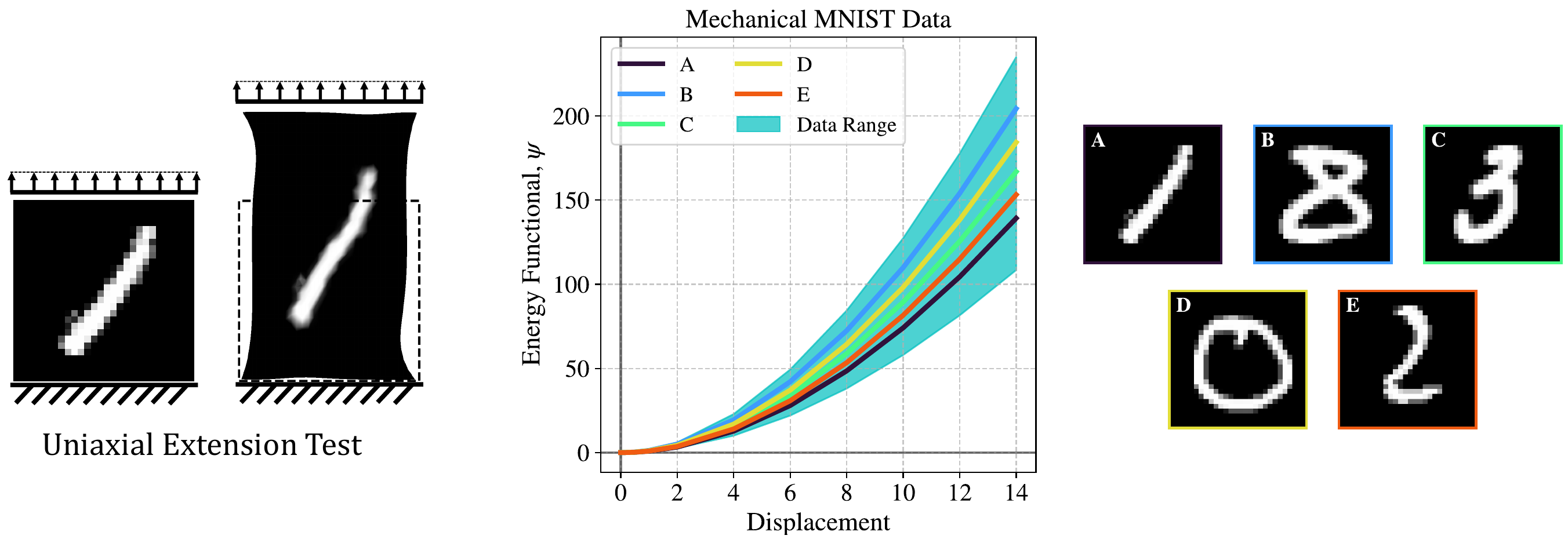} 
\caption{The Mechanical MNIST data set includes 60,000 training and 10,000 testing microstructures based on the MNIST handwritten digits data set and their corresponding total strain energy response curves in uniaxial extension.}
\label{fig:hyperelastic_database}
\end{figure}

The images are transformed into material properties using a compressible Neo-Hookean material model, where the strain energy functional $\psi$ is given by
\begin{equation}
\psi=\frac{1}{2} \mu[\mathbf{F}: \mathbf{F}-3-2 \ln (\operatorname{det} \mathbf{F})] \quad+\frac{1}{2} \lambda\left[\frac{1}{2}\left((\operatorname{det} \mathbf{F})^2-1\right)-\ln (\operatorname{det} \mathbf{F})\right],
\label{eq:mnist_neo}
\end{equation}
where $\tensor{F}$ is the deformation gradient and $\mu,\lambda$ are the Lamé coefficients.
The Lamé coefficients at every pixel are calculated through the corresponding Young's modulus:
\begin{equation}
E=\frac{\beta}{255.0}(100.0-1.0)+1.0 ,
\label{eq:mnist_young}
\end{equation}
where $\beta$ is the value of the grayscale bitmap and a constant Poisson's ratio of $\nu = 0.3$,  such that:
\begin{equation}
\lambda = \frac{E \nu}{(1+\nu)(1-2 \nu)} \quad\text{and} \quad \mu = \frac{E}{2(1+\nu)}.
\label{eq:lame}
\end{equation}
Thus,  the resulting microstructures in the data set have stiff elastic inclusions in a two orders of magnitude softer elastic domain.

Finite element simulations are then run on the resulting material domains,  with the bottom of the domain fixed and the top displaced according to a set of prescribed displacements. 
The applied displacements range from 0 to half the initial size of the domain. 
The finite element simulations are run using the FEniCS platform \citep{alnaes2015fenics} with a mesh size of 39,200 quadratic triangular elements which corresponds to approximately 140 elements per pixel in the MNIST image. The Mechanical MNIST data set includes load cases in addition to the uniaxial extension used in this work, such as shear, equibiaxial extension, and confined compression.
Fig.\ref{fig:hyperelastic_database} shows 5 samples from the training set as well as the error range of the responses in the entire Mechanical MNIST data set.
}

\bibliographystyle{plainnat}
\bibliography{main}

\end{document}